\documentclass[letterpaper]{article} 
\usepackage{aaai24}  
\usepackage{times}  
\usepackage{helvet}  
\usepackage{courier}  
\usepackage[hyphens]{url}  
\usepackage{graphicx} 
\def\UrlFont{\rm}  
\usepackage{natbib}  
\usepackage{caption} 
\usepackage{algorithm}
\usepackage{algorithmic}
\usepackage{bm}
\usepackage{multirow}
\usepackage{booktabs}
\usepackage{amsmath}
\usepackage{amsfonts}
\usepackage{algorithm}
\usepackage{algorithmic}
\usepackage{xspace}
\usepackage{cleveref}
\crefname{equation}{Eq.}{Eqs.}
\crefname{table}{Tab.}{Tables}
\crefname{figure}{Fig.}{Figures}
\crefname{algorithm}{Algorithm}{Algorithms}
\crefname{algocf}{Algorithm}{Algorithms}
\crefname{section}{Sec.}{Secs.}
\usepackage{tabularx}
\newcommand{\ourmethod}{BitCoin\xspace}

\newcommand{\etc}{\textit{etc.}}

\usepackage{newfloat}
\usepackage{listings}
\DeclareCaptionStyle{ruled}{labelfont=normalfont,labelsep=colon,strut=off} 
\floatstyle{ruled}
\newfloat{listing}{tb}{lst}{}
\floatname{listing}{Listing}
\title{\ourmethod: Bidirectional Tagging and Supervised Contrastive Learning based Joint Relational Triple Extraction Framework}
\author {
    Luyao He\textsuperscript{\rm 1},
    Zhongbao Zhang\textsuperscript{\rm 1},
    Sen Su\textsuperscript{\rm 1},
    Yuxin Chen\textsuperscript{\rm 1}
}
\affiliations {
    \textsuperscript{\rm 1}Beijing University of Post and Telecommunications\\
    heluyao@bupt.edu.cn, zhongbaozb@bupt.edu.cn, susen@bupt.edu.cn, chenyuxin@bupt.edu.cn
}

\begin{document}

\maketitle
\begin{abstract}
Relation triple extraction (RTE) is an essential task in information extraction and knowledge graph construction. Despite recent advancements, existing methods still exhibit certain limitations. They just employ generalized pre-trained models and do not consider the specificity of RTE tasks. Moreover, existing tagging-based approaches typically decompose the RTE task into two subtasks, initially identifying subjects and subsequently identifying objects and relations. 
They solely focus on extracting relational triples from subject$\rightarrow$object, neglecting that once the extraction of a subject fails, it fails in extracting all triples associated with that subject. 
To address these issues, we propose \textbf{BitCoin}, an innovative Bidirectional tagging and supervised Contrastive learning based joint relational triple extraction framework. Specifically, we design a supervised contrastive learning method that considers multiple positives per anchor rather than restricting it to just one positive. Furthermore, a penalty term is introduced to prevent excessive similarity between the subject and object. Our framework implements taggers in two directions, enabling triples extraction from subject$\rightarrow$object and object$\rightarrow$subject. Experimental results show that \ourmethod achieves state-of-the-art results on the benchmark datasets and significantly improves the F1 score on Normal, SEO, EPO, and multiple relation extraction tasks.
\end{abstract}

\section{Introduction}
Relation triple extraction plays a crucial role in information extraction and knowledge graph construction, and it aims to extract entity pairs and their corresponding relations in the form of (subject, relation, object) from unstructured text. Early RTE approaches mainly relied on pipelined methods, which involve a two-step paradigm: extracting entities through named entity recognition (NER) and then identifying relationships through relationship classification \cite{zelenko2003kernel,zhou2005exploring,chan2011exploiting}. Despite its flexibility, this paradigm falls short in overlooking the correlation and interaction between subtasks and potential error propagation. To address these challenges, researchers have proposed end-to-end approaches that simultaneously model entity recognition and relation classification, which have achieved improved RTE performance \cite{yu2010jointly,wei2019novel,eberts2019span,wang2020tplinker}.

Overlapping triples present a challenge where a single entity or pair of entities participating in multiple relational triples in the same sentence \cite{zeng2018extracting}. Among these existing approaches, tagging-based joint extraction methods have shown superior performance and capability in handling overlapping triples \cite{zheng2017joint,wei2019novel, yu2019joint, ren2022simple, zheng2021prgc}. 

Despite the promising results achieved by existing joint extraction methods, they still suffer from two major shortcomings. 
Firstly, existing joint extraction methods just employ generalized pre-trained models and do not consider the specificity of RTE tasks and the importance of designing additional tasks, resulting in unreliable extraction results. 
Secondly, existing tagging-based approaches typically decompose RTE tasks into two subtasks, which identify subjects initially and subsequently identify objects and relations. 
They solely focus on extracting relational triples from subject$\rightarrow$object, neglecting that once the extraction of a subject fails, it fails in extracting all triples associated with that subject.
These two issues significantly hamper the performance of RTE.

To overcome the limitations of existing approaches and achieve improved results, we propose \ourmethod, an end-to-end framework based on \textbf{\underline{B}}\textbf{\underline{i}}directional \textbf{\underline{t}}agging and supervised \textbf{\underline{Co}}ntrastive learn\textbf{\underline{in}}g. The core idea is first to employ contrastive learning method that considers multiple positives per anchor and designs a penalty term to prevent excessive similarity between subject and object. Then, we employ taggers in two directions, enabling the extraction of triples from subject$\rightarrow$object and object$\rightarrow$subject. 
There are two main challenges here.
Firstly, we aim to design a task to get better features by making related entities closer and unrelated entities farther apart, and one way to achieve this is through contrastive learning. However, existing contrastive learning methods are not applicable in this scenario. 
The conventional InfoNCE loss is unsuitable for cases where multiple positives may exist, as one subject can have relationships with multiple objects.
Also, the subject and object features may become excessively similar, making it difficult to distinguish them accurately, which results in incorrect triple extractions. 
Secondly, bidirectional extraction is not just about designing extractions in two directions but also about the interaction of information during the extraction in these two directions.

To address the first challenge, we propose a novel supervised contrastive learning method. 
Firstly, we consider multiple positives per anchor rather than limiting ourselves to just one positive. 
These positives are selected from samples of the same class, in contrast to the data augmentation approach commonly used in self-supervised learning. 
We designate a subject as an anchor, consider all relevant objects as positives, and then designate all non-relevant entities as negatives. To augment the number of positives and negatives, we employ dropout as a minimal form of data augmentation. 
Secondly, we design a penalty term to prevent excessive similarity between subject and object. When the similarity between subject and object reaches a threshold, this penalty term prevents its similarity from expanding further and stabilizes the similarity around the threshold.

To address the second challenge, we design taggers in two directions, enabling the extraction of triples from subject$\rightarrow$object and object$\rightarrow$subject. During extraction, information from both directions can interact with each other. For example, when extracting objects in the s2o direction, we can draw on the information of objects in the o2s direction for more accurate extraction. The extracted triples can be cross-validated by the results of two-way taggers, which means extractions from two directions complement each other. 
Also, we consider the fundamental properties of a triple, which are the interdependency and indivisibility of its entity pairs and relations. We may get unreliable results if we fail to utilize the relational information when extracting relevant objects fully but rely solely on the subject information to extract objects.
To adequately consider the information conveyed by the relation, we design a relationship prediction module to get relational features and combine them with sentence and entity features to get reliable triples. This comprehensive input integration enables us to fully exploit the relational information within the sentence, facilitating easier and more accurate triple extraction.
                                                         
Experimental results show that \ourmethod achieves state-of-the-art results on the benchmark datasets and significantly improves the F1 score on Normal, SEO, EPO, and multiple relation extraction tasks.

The main contributions of this work are as follows:
\begin{itemize}
    \item This is the first attempt to employ supervised contrastive learning with a penalty term for RTE tasks.
    \item We propose a novel end-to-end bidirectional tagging and supervised contrastive learning based framework \ourmethod. It significantly alleviates the problems of one-direction subject extraction failure and neglecting relationship information. 
    \item We evaluate our model on four public datasets, and the results indicate that our method outperforms all the state-of-the-art baselines, especially in complex scenarios.
\end{itemize}

\section{Related Works}
Researchers have proposed two kinds of methods for RTE: pipeline and joint learning. Traditional pipelined approaches initially perform named entity recognition to extract entities, followed by identifying relationships through relationship classification \cite{zelenko2003kernel,zhou2005exploring,chan2011exploiting}. 
Although flexible, these methods neglect the interdependence between the two subtasks and are prone to error propagation, consequently undermining overall performance.

To tackle this problem, several joint models that extract entities and relations jointly have been proposed. 
Among them, feature-based joint models require a process of constructing features manually. Neural network-based models have shown considerable improvement in both performance and efficacy in RTE tasks by replacing manual feature construction, which is a complicated process of feature-based joint models.
\citeauthor{zheng2017joint} \shortcite{zheng2017joint} proposed a novel tagging schema that unified the role of the entity and the relation between entities and converted the task of RTE to an end-to-end sequence tagging problem. However, they ignored the problem of overlapping triples.

To address this problem, various neural network-based joint extraction methods are proposed. \citeauthor{zeng2018extracting} \shortcite{zeng2018extracting} proposed a sequence-to-sequence model with copy mechanism to address the overlapping triples problem, although it struggles to generate multi-word entities. \citeauthor{fu2019graphrel} \shortcite{fu2019graphrel} also delved into the issue and devised a novel method based on Graph Convolutional Networks (GCNs).
As an improvement, \citeauthor{nayak2020effective} \shortcite{nayak2020effective} adopted an encoder-decoder model, where the decoder incrementally extracts words, just like machine translation models. 
\citeauthor{ye2021contrastive} \shortcite{ye2021contrastive} propose a contrastive triple extraction method with a generative transformer. \citeauthor{wang2020tplinker} \shortcite{wang2020tplinker} treated entity recognition and relation classification as a table-filling problem, where each entry represents the interaction between two words. \citeauthor{shang2022onerel} \shortcite{shang2022onerel} proposed to frame the joint extraction task as a fine-grained triple classification problem, which can extract triples from sentences in a one-module one-step manner.

Among these approaches, tagging-based joint extraction methods have shown superior performance and capability in handling overlapping triples. \citeauthor{wei2019novel} \shortcite{wei2019novel} proposed a novel cascade binary tagging framework that models relations as functions that map subjects to objects rather than treating relations as discrete labels on entity pairs, achieving competitive results. It first identifies all possible subjects and then identifies the relevant objects under all relations for each subject. \citeauthor{zheng2021prgc} \shortcite{zheng2021prgc} proposed an end-to-end framework that decomposed joint extraction into three subtasks: relationship determination, entity extraction, and subject-object alignment. Experiments show that tagging-based joint extraction methods achieve competitive results and have a solid ability to extract triples from sentences that contain overlapping triples or multiple triples.

\begin{figure*}[ht]
    \centering
    \includegraphics[width=\linewidth]{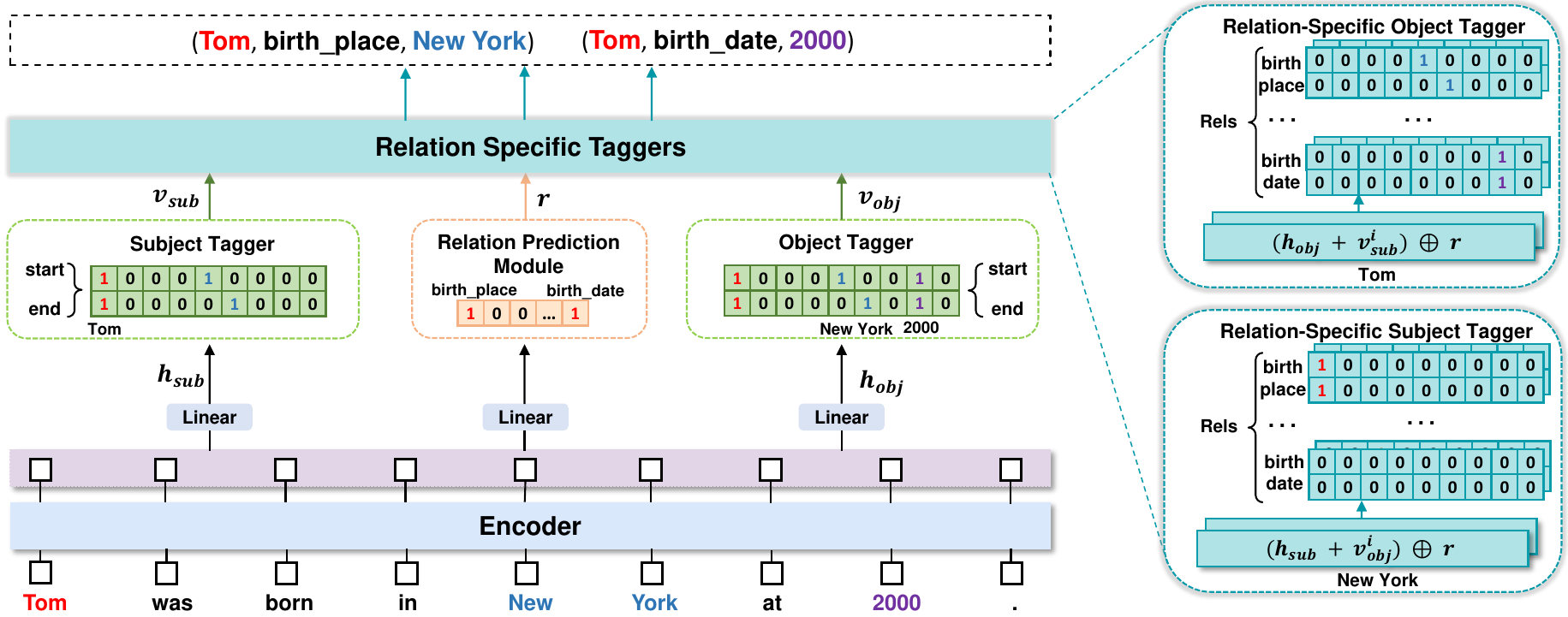}
    \caption{The overview structure of \ourmethod. In this example, given a sentence, \ourmethod detects two candidate subjects, three candidate objects and two potential relations. Then for each candidate subject, the relation-specific object tagger will extract relevant objects under all relations. The relation-specific subject tagger extracts relevant subjects under all relations for each candidate object. Finally, we take the union set of two-direction results.}
    \label{fig:model}
\end{figure*}

\section{Method}
The architecture of \ourmethod is shown in \cref{fig:model}. It consists of the following six components: encoder based on BERT and supervised contrastive learning, subject tagger, object tagger, relationship prediction module, relation-specific object tagger and relation-specific subject tagger. During training, we adopt a multi-task learning approach that allows each module to be trained with ground truth, resulting in a more reliable model. During the inference stage, \ourmethod operates in three stages: 1) The encoder generates representations for each token. 2) The subjects, objects and potential relationships are identified through the subject tagger, object tagger, and relationship prediction module. 3) The relation-specific object and subject taggers are used to tag the relevant objects and subjects under all relationships based on extracted subjects, objects and potential relations.

\subsection{Contrastive Learning based Encoder}
The encoder module is designed to convert sentences into word embeddings. Here, we first encode the input sentence using a pre-trained BERT model \cite{devlin2018bert} to generate the representations of each token in the sentence and then use our contrastive learning method to learn better features. We use 
the BERT-BASE-CASED model to ensure a fair comparison with other models, but it is theoretically possible to use other pre-trained models, such as Roberta and BART, \etc

In the knowledge graph, the ideal situation is that entities that are related are close to each other and entities that are not related are far apart. One way to achieve this is through contrastive learning. However, existing self-supervised contrastive learning methods are not applicable in this scenario, as one entity may have relations with multiple entities. Also, if we adopt the previous idea of comparative learning that pulls in the positives and pushes out the negatives, the subject and object features may become excessively similar, making it difficult to distinguish them accurately, resulting in incorrect triple extractions. 

To address these problems and obtain better features from the encoder for the RTE task, we have designed a supervised contrastive learning method with a penalty term. 
After generating the token representations through BERT, we use this contrastive learning method to train the BERT model further to obtain an encoder better suited for the RTE task.

\begin{table}[ht]
	\centering
	\begin{tabularx}{0.46\textwidth}{l|cc}
		\toprule
		\multirow{2}{*}{Input} 
        & \multicolumn{2}{l}{Tom was born in New York at 2000.} \\ 
        & \multicolumn{2}{l}{London is the capital of England.} \\
        \midrule
        \multirow{4}{*}{Triples} 
        & \multicolumn{2}{c}{(Tom, birth\_place, New York)} \\
        & \multicolumn{2}{c}{(Tom, birth\_date, 2000)}  \\
        & \multicolumn{2}{c}{(London, capital\_of, England)} \\
        & \multicolumn{2}{c}{(London, belong\_to, England)} \\
        \midrule
        anchor &
        Tom & 
        London\\
		\midrule
		Positives &
        New York, 2000 &
        England\\
		\midrule
        Negatives &
        London, England &
        New York, 2000\\
		\bottomrule
	\end{tabularx}
	\caption{Example of positives and negatives in supervised contrastive learning.}
 	\label{tbs:contrastive_learning}
\end{table}%


The idea of our supervised contrastive learning method is easy to understand. Firstly, we design a new loss function, as the self-supervised contrastive loss is incapable of handling the case where more than one sample is known to belong to the same class. 
Here, we set a subject as an anchor, all relevant objects as positives, and all non-relevant entities as negatives. \cref{tbs:contrastive_learning} shows an example of positives and negatives.
To increase the negatives and positives, we use dropout noise as the minimum form of data augmentation \cite{gao2021simcse} by inputting the sentence to the encoder twice and getting the embeddings of subjects and objects with different dropout masks.
Our loss function is structurally similar to the self-supervised contrastive learning loss, and the basic idea is both to pull in the positives and push out the negatives.
Differently, we consider many positives per anchor instead of only one positive, which can handle the situation where one subject may have multiple relevant objects. 
The supervised contrastive loss is computed based on the embeddings of subjects and objects, and the loss function is shown in \cref{con_l1.1}.

\begin{equation}
 \mathcal{L}_{1}^{i} = \frac{-1}{|P(i)|}\log \frac{\sum\nolimits_{p=1}^{P(i)} {e^{sim\left(\bm{h_i},\bm{h_i^p}\right)/\tau}}}{\sum\nolimits_{a=1}^{A(i)} e^{sim\left(\bm{h_i},\bm{h_i^a}\right)/\tau}},
  \label{con_l1.1}
\end{equation}

\noindent where $P(i)$ is the set of all positives, and $|P(i)|$ is its cardinality. $A(i)$ is the set of all negtives and positives, $h_i$ is the representation of the anchor, $\tau$ is the temperature hyperparameter and  $sim(\bm{h_1}, \bm{h_2})$ is the cosine similarity $\frac{\bm{h_1^T}\bm{h_2}}{{\Vert\bm{h_1}\Vert}\cdot{\Vert\bm{h_2}\Vert}}$.

Secondly, supposing we only use \cref{con_l1.1} as our loss function, we will keep drawing the subject closer to the relevant objects during training, making distinguishing the subjects and objects hard, resulting in incorrect triples. To solve this problem, we design a penalty term to prevent excessive similarity between the subject and object. When the similarity between the subject and the object reaches a threshold, this penalty term prevents its similarity from expanding further and stabilizes the similarity around the threshold. The penalty term is obtained by multiplying the following two terms, where the first term determines the strength of the penalty and the second term determines how the penalty is calculated. For the first term, we compute the difference between the similarity of the subject-object and threshold: the more the difference, the stronger the penalty. For the second term, it is structurally similar to the \cref{con_l1.1}, but the idea is changed to pull in the non-relevant objects and push out the relevant objects.
The penalty term is shown in \cref{con_l2}, and the total loss function for supervised contrastive learning is shown in \cref{con_l}.

 \begin{equation}
 \begin{aligned}
     \mathcal{L}_{2}^{i} = \bigg{[}-&\Big{(}\frac{1}{|P(i)|}\sum\limits_{p=1}^{P(i)}sim\left(\bm{h_i},\bm{h_i^p}\right)-\beta\Big{)} \\
     &\times \frac{1}{|N(i)|}\log \frac{\sum\nolimits_{n=1}^{N(i)} {e^{sim\left(\bm{h_i},\bm{h_i^n}\right)/\tau}}}{\sum\nolimits_{a=1}^{A(i)} e^{sim\left(\bm{h_i},\bm{h_i^a}\right)/\tau}}\bigg{]}_+
     \label{con_l2},
\end{aligned}
 \end{equation}

 \begin{equation}
     \mathcal{L}_{c}=\sum\limits_{i=1}^{I} \mathcal{L}_{c}^{i} = \sum\limits_{i=1}^{I}\left(\omega_1 \mathcal{L}_{1}^{i} + \omega_2 \mathcal{L}_{2}^{i}\right)
     \label{con_l},
 \end{equation}

\noindent where $N(i)$ is the set of all positives, $|N(i)|$ is its cardinality and $\beta$ is the threshold. 
 
\subsection{Bidirectional Tagging based Decoder}
\ourmethod is a bi-directional framework that extracts the relational triples in two directions: (1) extracting the subject, followed by both the relation and the object (s2o); and (2) extracting the object, followed by both the relation and the subject (o2s). The model structure is identical in both directions, and we will only present direction from subject$\rightarrow$object here.

\paragraph{\textbf{Subject Tagger}}
The Subject Tagger extracts all subjects by decoding h obtained from the encoder. Since the subjects, relations and objects in a triple have unique characteristics, we do not use the same features $h$ like CasRel, TPLinker and PRGC but use different features for each. Here, after obtaining the representations of each token from the input sentences, we can get the subject-specific representations through a fully-connected layer, denoted as 
$\bm{h}_{ {sub }}^{{i}}$, which is computed with \cref{sub_feature}.

\begin{equation} 
\bm{h}_{ {sub }}^{{i}}=\bm{W}_{ {sub}} \bm{h}^{{i}}+\bm{b}_{{sub}}
\label{sub_feature}
\end{equation}
where ${\bm{W}_{sub} \in {\mathbb{R}^{{d_h}\times{d_h}}}}$ is a trainable matrix, ${\bm{b}_{sub}\in\mathbb{R}^{{d}_{h}}}$ is a bias vector and ${{\bm{d}_{h}}}$ is the dimension of ${{\bm{h}_{i}}}$.

The Subject Tagger comprises two binary tagging modules that use a simple $(0/1)$ token to indicate whether the token is the start or end position of the subject. For each token, two separate probability formulas are used to calculate the probability that it is the start or end position of the subject. If the probability is higher than a threshold, the corresponding token position is tagged as $1$, indicating that the token is the start position of the subject. If it is lower, it is tagged as $0$. 
As there may be more than one subject in a sentence, we consider the closest starting and ending positions as one subject.
The probabilities of the start and end positions of the subject are calculated by \cref{sub1_prob,sub2_prob}.

\begin{align}
{p}_{ {sub\_start }}^{{i}}&={\sigma}\left(\bm{W}_{ {sub\_start }} \bm{h}_{{sub }}^{{i}}+\bm{b}_{{sub\_start }}\right)
\label{sub1_prob}
\\
{p}_{{sub\_end }}^{{i}}&={\sigma}\left(\bm{W}_{{sub\_end }} \bm{h}_{{sub }}^{{i}}+\bm{b}_{{sub\_end }}\right)
\label{sub2_prob}
\end{align}

\noindent where ${{{p}_{sub\_start}^{i}}}$ and ${{{p}_{sub\_end}^{i}}}$ represent the probabilities of identifying the $i$-th token in the input sequence as the start and end position of a subject respectively. 
${{\bm{h}_{sub}^{i}}}$ is the encoded representation of the $i$-th token with subject feature in the input sequence
and ${\sigma}$ is the sigmoid activation function.

\paragraph{\textbf{Relation Prediction Module}}
To obtain the relation-specific representations, we pass the representation of each token obtained from the encoder through a fully-connected layer, as shown in \cref{rel_feature}.
\begin{equation} 
\bm{h}_{ {rel }}^{{i}}=\bm{W}_{{rel}} \bm{h}^{{i}}+\bm{b}_{{rel}}
\label{rel_feature}
\end{equation}
where ${\bm{W}_{rel} \in {\mathbb{R}^{{d_h}\times{d_h}}}}$ is a trainable matrix, ${\bm{b}_{rel}\in\mathbb{R}^{{d}_{h}}}$ is a bias vector and ${{\bm{d}_{h}}}$ is the dimension of ${{\bm{h}_{i}}}$.

After obtaining the relation-specific representations, the Relation Prediction Module performs an Avgpool operation on these representations to obtain the representation of the input sentence. This representation is then used to extract all possible relations. Precisely, we can calculate the probability of each relation being present using a probability formula based on the average pooled representation of the input sentence. The corresponding relation is tagged as $1$ if the probability is higher than a threshold. The potential relations in the input sentence are obtained from \cref{rel_prob}.

\begin{align}
\bm{h}_{ {rel }}^{\operatorname{avg}}&={\operatorname{Avgpool}}\left(\bm{h}_{ {rel}}\right)\in{\mathbb{R}^{{d_h}\times{1}}}
\label{rel_avg}
\\
{p}_{rel}&={\sigma}\left(\bm{W}_{rel} \bm{h}_{\ {rel}}^{avg}+\bm{b}_{rel}\right)
\label{rel_prob}
\end{align}

\noindent where ${p}_{rel}$ represents the probability of identifying the relations in the input sequence and
$\bm{d}_{r}$ is the number of relations.

\paragraph{\textbf{Relation-Specific Object Tagger}}
The Relation-Specific Object Tagger extracts objects under all relations. Unlike the Subject Tagger, which uses encoder output directly, Relation-Specific Object Tagger also considers the subject tagged by the subject tagger, the relations predicted by the relation prediction module and object-specific representations from the object tagger in the other direction. This allows us to extract relations and objects simultaneously, which can solve the EPO and SEO problems. The module consists of $2\times {num}_{rels}$ binary tagging modules. Similar to the subject tagger, we calculate the probability that it is the start or end position of the corresponding object under each relation using two probability formulas, respectively. If the probability exceeds a threshold, the position corresponding to that token is tagged as $1$; otherwise, it is tagged as $0$. 
The probabilities of the start and end positions of the object under each relation are calculated by \cref{obj1_prob,obj2_prob}.

\begin{align}
\bm{h}_{rel\_obj}^i&=\bm{W}_{obj}\left(\left(\bm{h}_{obj}^i + \bm{v}_{sub}^{n}\right) \oplus \bm{p}_{rel}\right)+ \bm{b}_{obj} 
\label{rel_obj_feature}
\\
{p}_{obj\_start}^i&={\sigma}\left(\bm{W}_{obj\_start}^r\left(\bm{h}_{rel\_obj}^i\right) + \bm{b}_{obj\_start}^r\right) 
\label{obj1_prob}
\\
{p}_{obj\_end}^i&={\sigma}\left(\bm{W}_{obj\_end}^r\left(\bm{h}_{rel\_obj}^i\right) + \bm{b}_{obj\_end}^r\right)
\label{obj2_prob}
\end{align}

\noindent where ${{{p}_{obj\_start}^{i}}}$ and ${{{p}_{obj\_end}^{i}}}$ represent the probabilities of identifying the 
$i$-th token as the start and end position of an object respectively. ${{\bm{h}_{obj}^{i}}}$ is the representation of the $i$-th token with object feature, ${{\bm{v}_{sub}^{n}}}$ is the encoded representation of the $n$-th subject extracted in the subject tagger and ${\oplus}$ means concatenating two tensors. 

\begin{table*}[ht]
    \centering
    \resizebox{\textwidth}{!}{
        \begin{tabular}{l|cccc|ccccccccc}
            \toprule
            \multirow{2}{*}{Category} & \multicolumn{4}{c|}{Dataset} & \multicolumn{9}{c}{Details of Test Set} \\
            & Train & Valid & Test & Relation & Normal & SEO & EPO & N=1 & N=2 & N=3 & N=4 & N$\ge$5 & Triples
            \\
            \midrule
            \midrule
            NYT* & 56,195 & 4999 & 5000 & 24 & 3,266 & 1,297 & 978 & 3,244 & 1,045 & 312 & 291 & 108 & 8,110\\
            WebNLG* & 5,019 & 500 & 703 & 171 & 245 & 457 & 26 & 266 & 171 & 131 & 90 & 45 & 1,591\\
            NYT & 5,6195 & 5,000 & 5,000 & 24 & 3,222 & 1,273 & 969 & 3,240 & 1,047 & 314 & 290 & 109 & 8,120\\
            WebNLG & 5,019 & 500 & 703 & 216 & 239 & 448 & 6 & 256 & 175 & 138 & 93 & 41 & 1607\\
            \bottomrule
        \end{tabular}}
    \caption{Statistics of datasets. Note that a sentence can belong to both EPO class and SEO class.}
    \label{tbs:statistics_of_datasets}
\end{table*}

\subsection{Training Strategy}
All sub-modules of this framework work in a multi-task learning manner. In this way, each sub-modules have its loss function. The model is trained jointly by optimizing the combined objective function during training. The loss functions for the three modules mentioned above are denoted as ${{\mathcal{L}_{sub\_head}}}$, ${{\mathcal{L}_{sub\_tail}}}$, ${{\mathcal{L}_{rel}}}$, ${{\mathcal{L}_{rel\_obj\_head}}}$ and ${{\mathcal{L}_{rel\_obj\_tail}}}$, and they all utilize the binary cross-entropy loss function, as shown in \cref{BCE,sub_loss,obj_loss,rel_loss}.

\begin{align}
{\operatorname{BCE}}\left(\hat{y}, {y}\right)
&=-[{y}{log}\left({\hat{y}}) +(1-{y}\right)  \log \left(1 - {\hat{y}}\right)] 
\label{BCE}
\\
\mathcal{L}_{{s\_\left(h,t\right) }} 
&=\frac{1}{{l}} \sum_{{i}=1}^{l} {\operatorname{BCE}}\left({p}_{{s\_\left(h,t\right) }}^i, {{{y}}}_{ {s\_\left(h,t\right) }}^{{i}}\right)
\label{sub_loss}
\\
\mathcal{L}_{rel\_{o} \_(h, t)}
&=\frac{1}{{l}} \sum_{{i}=1}^l {\operatorname{BCE}}\left(p_{o\_{\left(h,t\right)}}^i, {{y}}_{{o} \_\left(h, t\right)}^{{i}}\right) 
\label{obj_loss}
\\
\mathcal{L}_{rel}
&=\frac{1}{r} \sum_{{i}=1}^r {\operatorname{BCE}}\left(p_{rel}^i, {{y}}_{rel}^{{i}}\right)
\label{rel_loss}
\end{align}

\noindent where ${\operatorname{BCE}}\left(\hat{y}, {y}\right)$ is a binary cross entropy loss,  $\hat{y}\in \left(0,1\right)$ is the calculated probability and y is the true label. l is the number of tokens in the input sentence, and r is the number of relations. 

Similar to the subject-to-object direction, there are four tagger losses in the object-to-subject direction, which are denoted as ${{\mathcal{L}_{obj\_head}}}$, ${{\mathcal{L}_{obj\_tail}}}$,  ${{\mathcal{L}_{rel\_sub\_head}}}$ and ${{\mathcal{L}_{rel\_sub\_tail}}}$, and are calculated by a formula similar to \cref{sub_loss,obj_loss,rel_loss}. The total loss is the sum of these parts, and it can be expressed by the following equation.

\begin{equation}
    \begin{aligned}
    \mathcal{L}_{total} = 
    \mathcal{L}_{\left(s, o\right)\_\left(h, t\right)} +
    \mathcal{L}_{rel\_\left(s, o\right)\_\left(h, t\right)} 
    +\mathcal{L}_{rel} + \mathcal{L}_{c}
    \label{loss}
    \end{aligned}
\end{equation}

\section{Experiment}
\subsection{Experiment Settings}
\paragraph{\textbf{Datasets}}
Following previous works \cite{wei2019novel,wang2020tplinker,zheng2021prgc,shang2022onerel}, we evaluate our framework on datasets NYT \cite{riedel2010modeling} and WebNLG \cite{gardent2017creating}. NYT and WebNLG have two different versions: one version only annotates the last word of entities, and the other annotates the whole span of entities. We denote the datasets based on the first standard as NYT* and WebNLG* and those based on the second standard as NYT and WebNLG. 

According to the different patterns of relational triple overlap, we classify the sentences into Normal, Entity-Pair-Overlap (EPO) and Single-Entity-overlap (SEO) classes to further study the capability of the proposed \ourmethod in handling complex scenarios. Detailed statistics of the two datasets are described in \cref{tbs:statistics_of_datasets}.

\begin{table*}[ht]
    \centering
    \resizebox{\textwidth}{!}{
        \begin{tabular}{lcccccccccccc}
            \toprule
            \multirow{3}{*}{Model} & \multicolumn{6}{c}{Partial Match} & \multicolumn{6}{c}{Exact Match} \\ 
            \cmidrule(r){2-7}
            \cmidrule(r){8-13}
            & \multicolumn{3}{c}{NYT*} & \multicolumn{3}{c}{WebNLG*} & \multicolumn{3}{c}{NYT} & \multicolumn{3}{c}{WebNLG} \\
            \cmidrule(r){2-4}
            \cmidrule(r){5-7}
            \cmidrule(r){8-10}
            \cmidrule(r){11-13}
            & Prec. & Rec. & F1 & Prec. & Rec. & F1 & Prec. & Rec. & F1 & Prec. & Rec. & F1 \\
            \midrule
            \midrule
            ETL-Span \cite{yu2019joint} & 84.9 & 72.3 & 78.1 & 84.0 & 91.5 & 87.6 & 85.5 & 71.7 & 78.0 & 84.3 & 82.0 & 83.1  \\
            CasRel \cite{wei2019novel} & 89.7 & 89.5 & 89.6 & 93.4 & 90.1 & 91.8  & - & - & - & - & - & -  \\
            RIN \cite{sun2020recurrent} & 87.2 & 87.3 & 87.3 & 87.6 & 87.0 & 87.3  & 83.9 & 85.5 & 84.7 & 77.3 & 76.8 & 77.0  \\
            TPLinker \cite{wang2020tplinker} & 91.3 & 92.5 & 91.9 & 91.8 & 92.0 & 91.9  & 91.4 & 92.6 & 92.0 & 88.9 & 84.5 & 86.7  \\
            SPAN \cite{sui2023joint} & 93.3 & 91.7 & 92.5 & 93.1 & 93.6 & 93.4  & 92.5 & 92.2 & 92.3 & - & - & -  \\
            CGT \cite{ye2021contrastive} & \textbf{94.7} & 84.2 & 89.1 & 92.9 & 75.6 & 83.4  & - & - & - & - & - & -  \\
            CasDE \cite{ma2021effective} & 90.2 & 90.9 & 90.5 & 90.3 & 91.5 & 90.9  & 89.9 & 91.4 & 90.6 & 88.0 & 88.9 & 88.4  \\
            RIFRE \cite{zhao2021representation} & \underline{93.6} & 90.5 & 92.0 & 93.3 & 92.0 & 92.6 & - & - & - & - & - & -  \\
            StereoRel \cite{tian2021stereorel} & 92.0 & 92.3 & 92.2 & 91.6 & 92.6 & 92.1  & 92.0 & 92.3 & 92.2 & - & - & -  \\
            PRGC \cite{zheng2021prgc} & 93.3 & 91.9 & 92.6 & 94.0 & 92.1 & 93.0  & 93.5 & 91.9 & 92.7 & 89.9 & 87.2 & 88.5  \\
            R-BPtrNet \cite{chen2021jointly} & 92.7 & 92.5 & 92.6 & 93.7 & 92.8 & 93.3  & - & - & - & - & - & -  \\
            BiRTE \cite{ren2022simple} & 92.2 & \textbf{93.8} & \underline{93.0} & 93.2 & 94.0 & 93.6  & 91.9 & \textbf{93.7} & 92.8 & 89.0 & 89.5 & 89.3  \\
            OneRel \cite{shang2022onerel} & 92.8 & 92.9 & 92.8 & \underline{94.1} & 94.4 & 94.3  & \textbf{93.2} & 92.6 & \textbf{92.9} & \underline{91.8} & \underline{90.3} & \underline{91.0}  \\
            \hline
            \hline
            \ourmethod(ours) & 92.9 & \underline{93.3} & \textbf{93.1} & \textbf{94.4} & \textbf{94.5} & \textbf{94.4} & \underline{93.1} & \underline{92.6} & \underline{92.8} & \textbf{91.9} & \textbf{90.5} & \textbf{91.2} \\
            \bottomrule
        \end{tabular}}
    \caption{Precision(\%), Recall (\%) and F1-score (\%) of our proposed \ourmethod and baselines.}
    \label{tbs:results}
\end{table*}

\begin{table*}[ht]
    \centering
    \resizebox{\textwidth}{!}{
        \begin{tabular}{lcccccccccccccccc}
            \toprule
            \multirow{2}{*}{Model} & \multicolumn{8}{c}{NYT*} & \multicolumn{8}{c}{WebNLG*} \\
            \cmidrule(r){2-9}
            \cmidrule(r){10-17}
            & Normal & SEO & EPO & N=1 & N=2 & N=3 & N=4 & N$\ge$5 
            & Normal & SEO & EPO & N=1 & N=2 & N=3 & N=4 & N$\ge$5
            \\
            \midrule
            \midrule
            CasRel & 87.3 & 91.4 & 92.0 & 88.2 & 90.3 & 91.9 & 94.2 & 83.7 
                   & 89.4 & 92.2 & 94.7 & 89.3 & 90.8 & 94.2 & 92.4 & 90.9\\
            TPLinker & 90.1 & 93.4 & 94.0 & 90.0 & 92.8 & 93.1 & 96.1 & 90.0
                     & 87.9 & 92.5 & 95.3 & 88.0 & 90.1 & 94.6 & 93.3 & 91.6\\
            SPN  & 90.8 & 94.0 & 94.1 & 90.9 & 93.4 & 94.2 & 95.5 & 90.6
                 & - & - & - & 89.5 & 91.3 & 96.4 & 94.7 & 93.8\\
            PRGC & 91.0 & 94.0 & 94.5 & 91.1 & 93.0 & 93.5 & 95.5 & 93.0
                 & 90.4 & 93.6 & 95.9 & 89.9 & 91.6 & 95.0 & 94.8 & 92.8\\
            OneRel & 90.6 & 94.8 & 95.1 & 90.5 & 93.4 & 93.9 & 96.5 & 94.2
                   & 91.9 & 94.7 & 95.4 & 91.4 & 93.0 & 95.9 & 95.7 & 94.5\\
            \midrule
            \midrule
            \ourmethod  & \textbf{91.1} & \textbf{95.0} & 94.9 & \textbf{91.2} & \textbf{93.6} & 93.8 &95.9 & \textbf{94.4}
                        & \textbf{92.0} & \textbf{94.9} & 95.7 & 91.1 & \textbf{93.4} & 95.7 & \textbf{95.7} & \textbf{94.7}\\
            \bottomrule
        \end{tabular}}
    \caption{F1-score (\%) on sentences with different overlapping patterns and different triple numbers}
    \label{tbs:overlapping}
\end{table*}
\paragraph{\textbf{Evaluation Metrics}}
The standard micro precision, recall, and F1 score are used to evaluate the results. There are two match standards for the RTE task: one is Partial Match that an extracted triple is regarded as correct if the predicted relation and the head of both the subject and object are correct, and the other is Exact Match that a triple is regarded as correct only when its entities and relationships are completely matched with a correct triple. We follow previous work \cite{wei2019novel,wang2020tplinker,zheng2021prgc,shang2022onerel} and use Partial Match on NYT* and WebNLG*, use Exact Match on NYT and WebNLG.

\paragraph{\textbf{Implementation Details}}
In our experiments, all training process is completed on a workstation with an Intel(R) 4210R 2.40G CPU, 128G memory, a single RTX 3090 GPU, and Ubuntu 20.04.
We use a small batch mechanism to train the model, with 4, 6, and 8 batch sizes. The learning rate is a linear warmup, the maximum learning rate is set to 1e-5, and the warmup step is set to the first quarter of the epoch. The threshold for judging whether there is a subject, an object, or a relation
is set to 0.5-0.6. The threshold for supervised contrastive learning in \cref{con_l2} is set to 0.85. The pre-trained BERT model is $[BERT- base, cases]$, which contains 12 Transformer blocks and 110M parameters, and the hidden size d is 768. All parameters are optimized by Adam algorithm \cite{kingma2014adam}. The dropout probability is 0.1. For a fair comparison, the maximum length of our model input sentences is set to 100 words, as suggested in previous works \cite{zeng2018extracting,fu2019graphrel}. 

We also use an early stopping mechanism to prevent the overfitting of the model. Specifically, we stop the training process when the performance on the validation set does not obtain any improvement for at least ten consecutive cycles. All involved hyperparameters are determined based on the results of the development set. Other parameters are initialized randomly. 

\paragraph{\textbf{Baselines}}
We compare \ourmethod with 13 strong baseline models and the state-of-the-art models ETL-Span \cite{yu2019joint}, CasRel \cite{wei2019novel}, RIN \cite{sun2020recurrent}, TPLinker \cite{wang2020tplinker}, CGT \cite{ye2021contrastive}, CasDE \cite{ma2021effective}, RIFRE \cite{zhao2021representation}, StereoRel \cite{tian2021stereorel},PRGC \cite{zheng2021prgc}, R-BPtrNet \cite{chen2021jointly}, BiRTE \cite{ren2022simple}, OneRel \cite{shang2022onerel} and SPAN \cite{sui2023joint}. Most results of these baselines are copied from their original papers directly.

\subsection{Experimental Results}

 \paragraph{\textbf{Overall Results}}
\cref{tbs:results} shows the comparison results of our model against 13 baselines on NYT and WebNLG in terms of Partial Match and Exact Match. Our \ourmethod method outperforms them in respect of almost all evaluation metrics. Especially on WebNLG*, \ourmethod obtains the best performance in terms of all three evaluation metrics. These results verify our motivation.

We attribute the outstanding performance of \ourmethod to its two advantages: 
Firstly, \ourmethod uses the novel supervised contrastive learning method with a penalty term to make the pre-trained model more suitable for the RTE task. We can get better features from the encoder through this method.
Secondly, we design taggers in two directions, and we can extract triples from subject$\rightarrow$object and from object$\rightarrow$subject. The two directions complement each other, and the extraction results can be cross-validated. 
Also, we notice the fundamental property of a triple, which is the interdependency and indivisibility of its entity pairs and relations. We designe the relation prediction model to fully exploit the relational information within the sentence, facilitating easier and more accurate triple extraction.

Compared with the tagging-based method CasRel which inspires us to treat relations as functions mapping subjects to objects instead of treating relations as discrete labels on entity pairs, \ourmethod achieves 3.5 and 2.6 absolute gain in F1-score on NYT* and WebNLG*, respectively. Such results confirm that contrastive learning and bidirectional framework are effective for RTE tasks.

\paragraph{\textbf{Detailed Results on Complex Scenarios}}
To verify the ability of our method on complex scenarios, we evaluate \ourmethod's ability for extracting triples from sentences that contain overlapping triples and multiple triples. This ability is widely discussed in existing models and is an important metric for evaluating the robustness of a model. For a fair comparison, we follow the settings of some previous models that classify sentences according to the degree of overlapping and the number of triples contained in a sentence and conduct two extended experiments on different subsets of NYT* and WebNLG*. We select five powerful models as baselines, and the detailed results are shown in \cref{tbs:overlapping}. 

It can be observed that \ourmethod has excellent superiority in handling complex sentences and achieves the best F1-score on 10 of the 16 subsets. Moreover, \ourmethod achieves more performance improvement when handling the sentences of the SEO class.
This is mainly because a single entity in an SEO sentence may be associated with multiple triples. Thus the existing models are more likely to suffer from the problem that once the extraction of an entity in some SEO triples is failed, all the associated triples of this entity would not be extracted either. However, the bidirectional framework in \ourmethod can effectively overcome such deficiency, and the mentioned issue almost does not affect it when handling the SEO sentences. This is also why \ourmethod performs well on sentences containing multiple triples.
In general, these two further experiments adequately show the advantages of our model in complex scenarios.

\paragraph{\textbf{Ablation Study}}
Here we make five kinds of detailed evaluations on \ourmethod, and the results are shown in \cref{tbs:xiaorong}. 
\begin{table}[ht]
    \centering
    \resizebox{0.405\textwidth}{!}{
        \begin{tabular}{l|lccc}
            \toprule
            \multicolumn{2}{c}{Model} & Prec. & Rec. & F1 \\ 
            \midrule
            \multirow{5}{*}{\rotatebox{90}{NYT*}} 
            & \textbf{\ourmethod} & \textbf{92.8} & \textbf{93.4} & \textbf{93.1} \\
            & - Contrastive Learning & 92.7 & 92.5 & 92.6 \\
            & - Direction from o2s & 92.3 & 92.2 & 92.4 \\
            & - Direction from s2o & 91.6 & 92.7 & 92.2 \\
            & - Relation Prediction & 92.5 & 92.3 & 92.4 \\
            \midrule
            \multirow{5}{*}{\rotatebox{90}{WebNLG*}} 
            & \textbf{\ourmethod} & \textbf{94.4} & \textbf{94.5} & \textbf{94.4} \\
            & - Contrastive Learning & 93.9 & 93.2 & 93.6 \\
            & - Direction from o2s & 93.7 & 92.3 & 93.0 \\
            & - Direction from s2o & 94.0 & 91.7 & 92.8 \\
            & - Relation Prediction & 94.1 & 94.1 & 94.1 \\
            \midrule
            \multirow{5}{*}{\rotatebox{90}{NYT}} 
            & \textbf{\ourmethod} & \textbf{93.1} & \textbf{92.6} & \textbf{92.8} \\
            & - Contrastive Learning & 92.9 & 92.1 & 92.5 \\
            & - Direction from o2s & 90.9 & 91.9 & 91.4 \\
            & - Direction from s2o & 91.6 & 91.3 & 91.5 \\
            & - Relation Prediction & 92.6 & 92.4 & 92.5 \\
            \midrule
            \multirow{5}{*}{\rotatebox{90}{WebNLG}} 
            & \textbf{\ourmethod} & \textbf{91.9} & \textbf{90.5} & \textbf{91.2} \\
            & - Contrastive Learning & 90.1 & 88.5 & 89.3 \\
            & - Direction from o2s & 90.6 & 87.8 & 89.1 \\
            & - Direction from s2o & 90.8 & 86.4 & 88.6 \\
            & - Relation Prediction & 91.7 & 87.5 & 89.5 \\
            \bottomrule
        \end{tabular}}
    \caption{Precision(\%), Recall (\%) and F1-score (\%) for ablation study of \ourmethod. }
    \label{tbs:xiaorong}
\end{table}

First, we evaluate the effectiveness of supervised contrastive learning. To this end, we implement a variant of \ourmethod without the supervised contrastive learning method. \cref{tbs:xiaorong} show that this performance drops on all datasets, indicating that our supervised contrastive learning is suitable for RTE tasks. We can get better word embeddings through this method.
Additionally, \cref{tbs:xiaorong} show that the performance of the one-directional tagging framework is much better than the CasRel model, which again indicates the effectiveness of our supervised contrastive learning method.

Second, we evaluate the effectiveness of the bidirectional tagging framework. To this end, we implement the following two variants of \ourmethod: (1) $\mathrm{\ourmethod_{s2o}}$, a variant that only uses the subject$\rightarrow$object direction to extract triples; (2) $\mathrm{\ourmethod_{o2s}}$, a variant that only uses the object$\rightarrow$subject direction to extract triples. \cref{tbs:xiaorong} show that the performance of both variants drops on all datasets, which shows the superiority of the proposed bidirectional tagging framework. Primarily, we take the union set of triples extracted from two directions as the final results. Both variants achieve lower perceptions and recalls, which indicates that the information from both directions can interact with each other, and the extracted triples can be cross-validated by the results of two-way taggers.

Third, we evaluate the effectiveness of the relation prediction module. To this end, we implement a variant of \ourmethod that neglect the potential relations obtained from the relationship prediction module. \cref{tbs:xiaorong} show that the performance drops on all datasets, which indicates that entity pairs and relations are interdependency and indivisibility, and we can obtain reliable results by fully utilizing the relational information present in the sentence during the extraction.

\section{Conclusion}
In this paper, we propose a novel bidirectional tagging and supervised contrastive learning based joint model named \ourmethod for RTE tasks. \ourmethod provides a general contrastive learning method that considers multiple positives per anchor and designs a penalty term to prevent excessive similarity between subject and object. Different from existing methods, taggers in our method are conducted in two directions, enabling the extraction of triples from subject$\rightarrow$object and object$\rightarrow$subject. Extensive experiments on four widely used datasets demonstrate the effectiveness of our method.

\appendix

\clearpage
\bibliography{aaai24}

\end{document}